\documentclass[10pt]{article}
\usepackage{geometry}
\usepackage{comment}
\usepackage{graphicx} 
\usepackage[frozencache,cachedir=.]{minted}
\usepackage{url}

\title{Evolutionary thoughts: integration of large language models and evolutionary algorithms}
\author{Antonio Jimeno Yepes \\
       Independent Researcher \\
       antonio.jimeno@gmail.com
       \and
       Pieter Barnard \\
       Independent Researcher \\
       pieterebarnard@gmail.com}

\date{}

\begin{document}

\maketitle

\begin{abstract}

Large Language Models (LLMs) have unveiled remarkable capabilities in understanding and generating both natural language and code, but LLM reasoning is prone to hallucination and struggle with complex, novel scenarios, often getting stuck on partial or incorrect solutions. However, the inherent ability of Evolutionary Algorithms (EAs) to explore extensive and complex search spaces makes them particularly effective in scenarios where traditional optimization methodologies may falter. However, EAs explore a vast search space when applied to complex problems.

To address the computational bottleneck of evaluating large populations, particularly crucial for complex evolutionary tasks, we introduce a highly efficient evaluation framework. This implementation maintains compatibility with existing primitive definitions, ensuring the generation of valid individuals.

Using LLMs, we propose an enhanced evolutionary search strategy that enables a more focused exploration of expansive solution spaces. LLMs facilitate the generation of superior candidate solutions, as evidenced by empirical results demonstrating their efficacy in producing improved outcomes.

The following GitHub repositories are available with this research:
\begin{itemize}
    \item Evolutionary algorithms with LLM support: \url{https://github.com/ajjimeno/llm-gp}
    \item Fast evolutionary evaluation: \url{https://github.com/ajjimeno/fast-evolutionary-evaluation}
    \item Experimental list data: \url{https://github.com/ajjimeno/list-data}
\end{itemize}

\end{abstract}

\section{Introduction}

Recent advances in the field of large language models (LLMs) have revealed remarkable capabilities to understand and generate both natural language and code~\cite{brown2020language, chowdhery2023palm, achiam2023gpt}. The recent explosion of LLM based reasoning research, Chain-of-Thought\cite{Wei2022Chain-of-Thought}, Tree-of-Thoughts\cite{Yao2023Tree}, among others, has fundamentally shifted what we expect from the field of artificial intelligence. These models, trained on vast datasets, possess the potential to infuse semantic understanding and more targeted generation into a wide array of computational tasks. Using the knowledge and pattern recognition capabilities of LLMs, it may become possible to change how the search is conducted in a solution space.
However, we must acknowledge the inherent limitations of these methods: LLM reasoning can be prone to hallucinations and struggle with complex, novel scenarios, often getting stuck on partial or incorrect solutions.

Evolutionary Algorithms (EAs), inspired by the principles of natural selection and genetics, stand as a powerful meta-heuristic search algorithms widely used to solve intricate optimization problems \cite{holland1992adaptation}. These algorithms operate on a population of candidate solutions, iteratively refining them through processes analogous to biological evolution, including fitness evaluation, selection of the fittest, and the application of genetic operators such as crossover and mutation \cite{mitchell1998introduction, eiben2015introduction}. The inherent ability of EAs to explore extensive and complex search spaces makes them particularly effective in scenarios where traditional optimization methodologies may fail \cite{holland1992adaptation, forrest1993genetic, goldberg1989genetic}. This exploration capability makes EAs a valuable tool for navigating the large function spaces often encountered in complex computational tasks, including the search for optimal or near-optimal programs.

EAs face several inherent challenges that limit their applicability and efficiency when faced with highly complex problems. These limitations include a potentially long convergence rate, the difficulty in effectively exploring exceptionally large search spaces, the persistent risk of premature convergence to suboptimal solutions, and the considerable computational resources required to evaluate large populations of candidate solutions \cite{deb2011multi, yao1999evolving, spears1993crossover, radcliffe1993genetic, eshelman1991preventing, mahfoud1995niching}. The computational expense is particularly pronounced when evaluating the fitness of a large number of individuals, a necessity for maintaining diversity and exploring broad solution spaces, especially in tasks involving the evolution of complex programs where fitness evaluation might involve resource-intensive compilation and execution \cite{langdon1996genetic, koza1993genetic, miller1989designing}.

In our research, we propose an integration of LLM-powered search with a highly efficient implementation for the evaluation of large populations. This hybrid methodology is designed to address the inherent limitations of traditional EAs by enabling a more intelligently guided search through the vast program space and by significantly accelerating the evaluation process, thereby facilitating the application of EAs to more complex and demanding problems. We propose two main contributions:

\begin{enumerate}
    \item Using Large Language Models (LLMs) capabilities, we propose an enhanced evolutionary search strategy that enables a more focused exploration of expansive solution spaces. By intelligently guiding the search, LLMs facilitate the generation of superior candidate solutions, as evidenced by empirical results demonstrating their efficacy in producing improved outcomes.
    \item To address the computational bottleneck of evaluating large populations, particularly crucial for complex evolutionary tasks, we introduce a highly efficient evaluation framework. This implementation maintains compatibility with existing primitive definitions, ensuring the generation of valid individuals. However, it significantly accelerates the compilation and execution phases through optimized individual representations and underlying computational strategies, enabling rapid fitness assessments for large populations. The implementation of the evaluation framework works on multiple CPU cores or on CUDA GPUs.
\end{enumerate}

This research is organized as follows; in the following section, related work, we review related work in EAs and their relation to large language models. Then, we present the methods and proposals section which looks at combining EAs and large language models. This is followed by the results and discussion section, and finally we present some concluding remarks on the results and future work.

\section{Related Work}

Although EAs have proven to be a powerful problem-solving paradigm, their application to highly complex problems often encounters significant difficulties. These challenges stem from the nature of the search space, the probabilistic operators employed, and the computational demands of the evolutionary process itself.

One of the primary hurdles in applying evolutionary algorithms to complex problems lies in navigating large search spaces~\cite{langdon1996genetic}. Traditional EAs rely on random mutation and crossover operations to explore this expansive space \cite{forrest1993genetic}. However, in extremely large search spaces, these purely probabilistic methods can become inefficient, potentially taking an inordinate amount of time to locate promising solutions or even failing to explore the most relevant regions effectively \cite{forrest1993genetic}.

Another significant limitation of traditional EAs is the problem of premature convergence~\cite{deb2011multi, yao1999evolving, spears1993crossover, radcliffe1993genetic, eshelman1991preventing, mahfoud1995niching}. This phenomenon occurs when the population of candidate solutions converges too quickly to a suboptimal solution, effectively getting stuck before the entire search space has been adequately explored \cite{yao1999evolving}.

The efficiency of fitness evaluation is a critical factor in the overall performance and scalability of EAs, particularly when dealing with large populations that are often necessary to address complex problems \cite{langdon1996genetic, koza1993genetic, miller1989designing}. Several techniques have been proposed and implemented to improve the speed of this process.

One fundamental aspect is the faster implementation and individual representation. Efficient coding practices and the use of optimized data structures to represent individuals and perform fitness calculations can significantly reduce the computational overhead associated with each evaluation.

The intersection of Large Language Models (LLMs) and evolutionary computation represents a new area of research with the potential to significantly improve optimization processes and problem solving strategies \cite{brown2020language, chowdhery2023palm, achiam2023gpt, chen2022program, austin2021program, nijkamp2022codegen, fried2022incoder, li2022pre, ziegler2019fine, madaan2023self, wei2022chain, kojima2022large, sanh2021multitask, thoppilan2022lamda, rae2021scaling, hoffmann2022training, touvron2023llama}.

Recent research has begun to explore the integration of Large Language Models (LLMs) to specifically guide the search process within EAs, aiming to overcome the limitations of purely random exploration in vast search spaces \cite{brown2020language, chowdhery2023palm, achiam2023gpt}. This field recognizes the potential of LLMs to inject a level of semantic understanding and strategic decision-making into the traditionally stochastic nature of EAs \cite{chen2022program, austin2021program, nijkamp2022codegen}. By leveraging LLMs to analyze the problem space, suggest promising mutations, and even dynamically adjust the fitness landscape, researchers are striving to create a more efficient and targeted evolutionary process \cite{li2022pre, ziegler2019fine, madaan2023self}. However, the inherent challenges of LLM hallucination and potential biases are also acknowledged, necessitating a careful balance between harnessing the power of language models and maintaining the robustness of evolutionary search \cite{wei2022chain, kojima2022large, sanh2021multitask}. The goal is to create a symbiotic relationship, where LLMs provide intelligent guidance and EAs offer the resilience and adaptability needed to navigate complex and dynamic problem domains \cite{thoppilan2022lamda, rae2021scaling, hoffmann2022training}. This exploration is not just about improving performance; it is about fundamentally rethinking how we approach complex optimization, blending the strengths of large language models with the power of evolutionary adaptation \cite{touvron2023llama}.

\section{Methods}

In this section, we describe the way in which the LLM contributes to the evolutionary algorithm.
We initially describe how an evolutionary algorithm works in general and then we provide specific decisions made for our research.
Whenever a large language model is used in conjunction with the evolutionary algorithm, it will be explained for each component of the evolutionary algorithm.

\subsection{Tasks definition}

We have designed a set of tasks with multiple levels of difficulty, each requiring the learning of a function that is evaluated on a set of test examples. The primary challenge lies in analyzing the training examples within each instance to identify the underlying transformation necessary to generate the correct output for the test cases.

Each training instance comprises a set of input-output pairs, where the input is a list of integers, and the expected output is another list of integers. Each integer falls within the range $[0,9]$ and the goal is to infer the function that maps inputs to outputs based on the observed patterns in the training data.
This task design is similar to the Abstract and Reasoning Challenge (ARC)~\cite{chollet2024arc}.

EAs are traditionally used for optimization problems, where candidate solutions evolve based on a predefined fitness function \cite{goldberg1989genetic, holland1992adaptation}. However, the described task deviates from conventional evolutionary algorithm applications as it focuses on learning a function that generalizes from training instances to unseen test data. This problem is more in line with inductive reasoning and program synthesis, where the objective is to infer rules from structured input-output mappings rather than directly optimizing numerical or combinatorial functions \cite{gulwani2015inductive, schmidt2009distilling}. 

A description of the tasks that we have considered in this work is shown below in ascending order of complexity.

\begin{enumerate}
    \item Count: the function has to output the number of elements in the input list. This task does not necessarily need the function to understand the training examples of the instance.
    \item Max-min: the function has to output the maximum or minimum of the input list. The function needs to understand from the training examples if the maximum or minimum needs to be returned considering the training examples of the instance.
    \item Inverse: the examples show a list in which the elements are inverted or the elements are in the same position as in the input list. The function needs to understand what type of processing is done on the training examples before deciding what to do with the testing output list.
    \item Sort: training examples show a list that is sorted in ascending or descending order. The function needs to understand what sorting is happening and apply it to the testing output list.
\end{enumerate}

An example is shown below (listing~\ref{lst:sort_task_example}) for the sort task.
In this example, the elements are sorted in descending order.

\begin{listing}[htpb!]
\begin{minted}[frame=lines, mathescape, fontsize=\footnotesize]{python}
Training

Example 1

Input: [5, 4, 3]
Output: [5, 4, 3]

Example 2

Input: [8, 5, 3, 9, 0, 0, 4, 4, 1, 6]
Output: [9, 8, 6, 5, 4, 4, 3, 1, 0, 0]

Testing

Input: [1, 3, 4, 0, 6, 0, 5, 4, 3]
Output: [6, 5, 4, 4, 3, 3, 1, 0, 0]
\end{minted}
\caption{Sorting task example. In this example, the input list is sorted in descending order in the output.}
\label{lst:sort_task_example}
\end{listing}

\subsection{Evolutionary algorithm}

In the initial step, $initial\_population\_generator$ generates $population\_size$ individuals.
Evolutionary operations are run over the population for $n\_generations$.
During each of these generations, the individuals with the highest fitness score are selected to proceed to subsequent steps in the evolution process. Then, depending on certain probability, these individuals of the population are mutated and crossover is applied to pairs of individuals. The result is a set of possible solutions evolved from the best candidates of the previous round.
The pseudocode for the evolutionary algorithm is presented in listing~\ref{lst:evolutionary_algorithm}.

\begin{listing}[htpb!]
\begin{minted}[frame=lines, mathescape, fontsize=\footnotesize]{python}

population = initial_population_generator(population_size)

for g in 1..n_generations:

  population = selection(population, population_size)

  for i in 1..population_size:
    if random_number > mut_prob:
        population[i] = mutation_function(population[i])

  for i in 1..population_size (step 2):
    if random_number > xover_prob:
        population[i], population[i+1] = xover_fuction(population[i],
                                                       population[i+1])

\end{minted}
\caption{Evolutionary algorithm}
\label{lst:evolutionary_algorithm}
\end{listing}

\subsubsection{Individuals in the population}
\label{subsec:population_individual}

In the context of this research, an individual is a program defined as a tree.
The nodes in the tree are primitives, which can be leaf or non-leave primitives.
The arguments and the return values of the primitives are typed, and leaf nodes have no arguments.

In this implementation, which relies on the DEAP framework \cite{de2012deap}, primitive types are defined as Python classes.
We have defined several types, which include different types of integers depending on if the integer is from the training (WInteger) or testing set (RInteger).
It can be noted in listing~\ref{lst:types-definition}, that the different types of integer are defined as a class tree.

\begin{listing}[htpb!]
\begin{minted}[frame=lines, mathescape, fontsize=\footnotesize]{python}
class Operation(object):
    pass

class COperation(Operation):
    pass

class Integer(object):
    pass

class WInteger(Integer):
    pass

class RInteger(Integer):
    pass

class Boolean(object):
    pass
\end{minted}
\caption{Types definition}
\label{lst:types-definition}
\end{listing}

Primitives are defined with a name and the input and output argument types, with examples available in listing~\ref{lst:primitive-definition}.

\begin{listing}[htpb!]
\begin{minted}[frame=lines, mathescape, fontsize=\footnotesize]{python}
pset.addPrimitive(r.testing_output_write, [RInteger], Operation)
pset.addPrimitive(r.testing_input_min, [], RInteger)
\end{minted}
\caption{Primitive declaration,~\textit{testing\_output\_write} is a writing operation on the testing output list with an integer from the testing lists as input and~\textit{testing\_input\_min} returns the minimum element of the testing input list.}
\label{lst:primitive-definition}
\end{listing}

Listing~\ref{lst:example-program} shows an example of a program with the leaf primitive $testing\_input\_min$, which returns the minimum value of the testing input list, and $testing\_output\_write$ that writes a value to the current position in the testing output list.

\begin{listing}[htpb!]
\begin{minted}[frame=lines, mathescape, fontsize=\footnotesize]{python}
testing_output_write(testing_input_min())
\end{minted}
\caption{Example program, which reads the minimum number from the input list of the testing example and writes it to the current position in the testing output list.}
\label{lst:example-program}
\end{listing}

\paragraph{Fitness function:}

In our task, the fitness metric is the average accuracy among all the examples in the ground truth.
The output is a list and the accuracy is calculated as the number of correct list elements vs. the total number of elements.
Elements in the list are ordered, so the position in the list is relevant.

While accuracy is the metric considered for fitness, function length, based on the number of primitives, is used to compare two functions with the same fitness value. Two programs could have the same accuracy metric when measured against the training set but have different generalization risks.
To minimize the generalization risk, we have considered the shortest length heuristic~\cite{vapnik1995nature,vapnik1998statistical}.
Function length is bounded to an upper function length using the length of the function with the best fitness and the shortest length with a slack of 25 primitives.
This is done before function selection, as described below.

\subsubsection{Initial population generation:}

Given the population size, the EAs generates randomly valid functions. The initial set of functions have bounded the number of primitives, which promotes searching for more complex functions starting from more simple ones.

With the proposed approach supported by an LLM, we obtain an initial set of individuals, which are informed by the task the algorithm should solve.
The set of individuals is limited to a maximum of 30 valid individuals generated by the LLM.

First, the LLM is shown examples of the task to be solved from the training set.
A set of 5 explanations are generated using the LLM.
The explanations are provided to an LLM with the training set instances and it is asked to use those explanations to provide a final description of what may be happening in the input examples.

Given the description of the problem generated by the LLM, the LLM is requested to generate functions that might solve the task.
The functions are evaluated so they are valid functions.
The LLM is called several times till at least 30 valid functions are generated.

\subsubsection{Selection}

The population size is a hyperparameter of EAs and determines the maximum number of individuals after each iteration.
During selection, the fitness of the new individuals is estimated and the population size is used to keep only the fittest individuals.
For the selection of individuals, we keep a population of a given size.
After the mutation and crossover operations, there are new individuals that are evaluated. 

\paragraph{Elitism}

It has been shown that keeping the fittest individual in the population helps with the convergence and stability of the evolutionary algorithm~\cite{de1975analysis,goldberg1989genetic}.
In our work, we keep the top individual identified so far and add it back to the population.

\subsubsection{Mutation}

For this type of program tree, a mutation is defined as selecting a random node in the tree and changing it for a new subtree which is randomly generated and makes a valid tree.
A mutation has only a certain probability of being applied to an individual of the population.

Mutations are random and no information about the task being solved is used.
We propose an update to the mutation operation by using an LLM on examples where a given individual might have failed at achieving the task. This information is available from the evaluation of each individual with respect to the task.

As in the generation of the initial individuals of the population, we consider only a limited number of individuals to be mutated using the LLM. The rest of the individuals follow the same mutation scheme as the baseline approach.
We observed that the LLM tends to produce mutations that reduce the diversity of the population. We have reduced the probability of individuals that are LLM mutated. These mutations are added to the population pool and do not substitute existing individuals.

We also guarantee that the selected elitist individual is mutated with the LLM and added to the population.
LLMs do not always succeed in the generation of syntactically correct functions and up to 3 retries are used before finally failing. 

\subsection{Crossover}

During crossover, two functions are selected given a probability. A random subtree is selected from the function of one parent and swapped with a randomly chosen subtree from the function of the other parent. This process enables the exchange of functional building blocks between solutions, promoting diversity and potentially leading to more effective programs.
We did not use an LLM to perform crossover, which could be explored as future work.

\subsection{Fast evaluation of evolutionary programs}

For complex problems, a large population size provides quite a variety of individuals, which increases the probability of finding an appropriate program.
Evaluation of a large number of programs is expensive.
Existing libraries such as DEAP~\cite{de2012deap,fortin2012deap}, use the Python interpreter as the compiler and evaluation of programs generated using Python functions, which allows for fast prototyping.

With a small population of individuals, speed is not a problem.
However, a larger population makes it really slow, the environment needs to be prepared, the program of the individual compiled and the fitness needs to be evaluated.
We propose an implementation of a fast evaluation algorithm implemented in C++\footnote{Fast evolutionary evaluation: \url{https://github.com/ajjimeno/fast-evolutionary-evaluation}}.
Also, memory requirements grow with population size.
In Python, pointers are expensive.
We have updated the DEAP trees with array pointers to entries in a table of node types.

There is an implementation that works with NVIDIA CUDA (Compute Unified Device Architecture)\footnote{\url{https://developer.nvidia.com/cuda-toolkit}}.
On our hardware, the CPU based implementation with multiple cores is faster than the CUDA based implementation.
The reason for the performance difference is due to warp divergence\footnote{CUDA Warp divergence:\url{https://developer.nvidia.com/blog/cooperative-groups}}.

\subsubsection{Primitives}

Individuals in the population are programs represented as trees in which the nodes are from a set of predefined primitives, as described in section~\ref{subsec:population_individual}. 
The primitives perform: operations on lists (read/write), boolean operations, or decide how to branch the tree processing.
A description of the primitives is available from~\url{https://github.com/ajjimeno/llm-gp/blob/78f279a9d8459d68e6dbf50789faedb9d4f160bd/prompts.py#L45}.

The primitives intend to run over the lists that represent the training and testing instances, which contain lists representing the input and output of the programs.

\subsubsection{Compilation and running of a function}

The C++ evaluation library can compile functions as strings or as lists of pointer trees, generated by the evolutionary algorithm.
Function compilation generates a set of nodes organized as a tree. Each node points to its arguments, if any. 
Trees are stored as a list of nodes and the lists of trees are stored in a continuous list in memory, which helps with faster transfer of this list to the GPU.
When the pointer is at a certain node, the node contains information about the type of node, which defines the function that needs to be run, and the arguments.

Each function tree runs a predefined number of iterations (e.g. 200).
In each iteration, the tree is traversed and only the nodes that comply with certain conditions are traversed.

Function call in CUDA is expensive, all registers need to be saved and recovered after the call. Initial implementation relying on a table of function pointers, prevents basic optimizations such as function inlining by the compiler.
To solve this issue, a stack and a switch-case are used to traverse the list of nodes. 
The root node is pushed to the stack and depending on the execution of the node, the following arguments might be loaded to the stack.
Once the stack is empty, the execution is finished.
There is a set of three variables used as registers that are used to store values from function calculation.

Example of the structure of the stack traversal and the switch-case are shown in listing~\ref{lst:switch-case}.
The $while$ runs till the stack pointer is at the top of the stack.
The~\textit{switch-case} runs each one of the node types defined as $case\_operation$. The following is an example for $get0$.

\begin{listing}[htpb!]
\begin{minted}[frame=lines, mathescape, fontsize=\footnotesize]{python}
while (s_pointer > 0)
{
    SNode *node = &stack[--s_pointer];

    switch (node->case_operation)
    {
    case 0:
        reg = get0(NULL);
        break;
    ...
    default:
        run->status = -1;
        break;
}
\end{minted}
\caption{Structure of the switch-case}
\label{lst:switch-case}
\end{listing}

The listing~\ref{lst:loop-function} shows the implementation of the loop function.
The fist use case 83, adds two entries in the stack, the argument that needs to be repeated n-times and the argument that provides the number of n repetitions.
We see that the n-times argument is stored in the~\textit{reg} registry.
As well, there is a limit to the nested loops, set to 3 and identified by the variable~\textit{run\texttt{->}inner\_loop} and the loop is indicated by a positive number less than or equal to 30.

\begin{listing}[htpb!]
\begin{minted}[frame=lines, mathescape, fontsize=\footnotesize]{python}
    case 83: // loop
    {
        run->inner_loop++;

        // Add operation node n-times
        Node *pnode = &run->nodes[node->node_pointer];

        stack[s_pointer++] = {pnode->args[1], 100};

        // Add read value node
        stack[s_pointer++] = {pnode->args[0], run->nodes[pnode->args[0]].pointer};
    }
    break;
    case 100:
        if (run->inner_loop < 3 && reg > 0 && reg <= 30)
        {
            //  Read register
            stack[s_pointer++] = {node->node_pointer, 110};
            for (int i = 0; i < reg; i++)
            {
                //  Loop operation
                stack[s_pointer++] = 
                        {node->node_pointer, 
                         run->nodes[node->node_pointer].pointer};
            }
        }
        else
        {
            run->status = -1;
        }

        break;
\end{minted}
\caption{Implementation of the loop function}
\label{lst:loop-function}
\end{listing}

\subsection{Evolutionary algorithm and large language models}

Generation of seed individuals.

\begin{listing}[htpb!]
\begin{minted}[frame=lines, mathescape, fontsize=\footnotesize]{python}
description = description_analyzer(
                    [generate_description(task_set)
                     for i in range(5)], task_set)

while len(population) < 30:
  population += 
    generate_individuals(
        description, primitives, task_set)
\end{minted}
\caption{Generation of seed individuals using an LLM}
\label{lst:llm-seed-generation}
\end{listing}

Mutation supported by a large language model.

\begin{listing}[htpb!]
\begin{minted}[frame=lines, mathescape, fontsize=\footnotesize]{python}
count = 0 # Keeps track of retries

while count < 3:
    try:
        new_program = 
            get_guided_mutation_program(task_description, individual, score)
        return new_program
    except Exception:
        count += 1
\end{minted}
\caption{Mutation generation using an LLM}
\label{lst:llm-mutation-generation}
\end{listing}

\section{Results}

\subsection{Datasets}

The results have been obtained synthetically, in which the input list has to be transformed according to an unknown function that needs to be predicted.
Each instance consists of training examples that can be used as reference and a testing example, on which the predicted function can be evaluated.
For each task, 100 examples for training and 100 examples for testing have been generated.

\subsection{Evolutionary algorithm}

We have run experiments with different hyperparameters, which include the number of examples used for training, the number of individuals in a population and the number of evolutionary steps.

Selection is based on Stochastic Universal Sampling~\cite{baker1987reducing}.
The selection is made by using a single random value to sample all
individuals by choosing them at evenly spaced intervals. The returned list
contains references to the input individuals.

We have updated the fitness prior to the following equation, which provides a way to bias the selection towards the fittest individuals, while still keeping less fit individuals.
\begin{equation}
\hat{fitness} = e^{fitness*50}-1
\end{equation}

Evaluation of the fitness of an individual is based on the percentage of the average of correct outputs. An individual could evolve endlessly with a large number of branches that are never traversed using the training data. We have used as reference the individual with the best accuracy and the shortest length as reference. At any given generation, no program should be longer than the selected best individual given a slack on the length. This slack is to ensure that there is exploration of potential individuals.
Also, we apply~\textit{elitism}, selecting the individual with the best fitness and the shortest length, which is passed to the next generation.

We considered filtering of programs by length because there are longer programs with no better performance, which add noise and might reduce the performance of examples in the testing set.
On the other hand, unless we have simple tasks, finding if a function found by our algorithm is the shortest is undecidable in the general case~\cite{chaitin1977algorithmic}, but we can try simplifying the functions found for a simple task and then measure the difference.

The evolutionary algorithm was run 10 times and the results show average metrics on the top individual of each run after 1500 generation steps.
We decided on a large number of generations for two reasons: we do not know in advance what might be the appropriate number of generations that might be needed and the performance of the top function on the training set will not reduce over time and, hopefully, with the search of shorter functions might not damage the performance on the testing set, which we are able to examine having a closer look at the results.
The training set for each task is used during the training phase and accuracy is estimated using the top individual from the last generation step.

Table~\ref{tab:evolutionary-algorithm-results} shows the results using the evolutionary algorithm with different population sizes.
We see that both~\textit{count} and~\textit{max-min} tasks obtain perfect performance with all population sizes and the length of the generated functions are shorter compared to the other tasks with lower height.
This shows that it is not challenging to find a function with high performance from the set of all of possible functions.
\textit{max-min} has a shorter top individual that appears with a larger population size.

Both~\textit{inverse} and~\textit{sorted} tasks get better average performance with larger population sizes.
The top individual's length and height decrease with larger population sizes.
Larger population sizes consume more memory and computing time, but it seems to help in the convergence of the evolutionary algorithm, which supports the implementation of the fast evaluation algorithm presented in the Methods section.

\begin{table}[htpb!]
\begin{small}
\begin{center}
\begin{tabular}{|c|r|r|r|r|r|}
\hline
 Task   & PS  & Training & Testing & Length & Height \\
\hline
count & 300 & 1.000 (0.00) & 1.000 (0.00) & 2.00 (0.00) & 1.00 (0.00) \\
count & 1000 & 1.000 (0.00) & 1.000 (0.00) & 2.00 (0.00) & 1.00 (0.00) \\
count & 3000 & 1.000 (0.00) & 1.000 (0.00) & 2.00 (0.00) & 1.00 (0.00) \\
count & 30000 & 1.000 (0.00) & 1.000 (0.00) & 2.00 (0.00) & 1.00 (0.00) \\
\hline
max-min & 300 & 1.000 (0.00) & 1.000 (0.00) & 10.00 (0.89) & 3.00 (0.45) \\
max-min & 1000 & 1.000 (0.00) & 1.000 (0.00) & 9.70 (1.27) & 2.80 (0.60) \\
max-min & 3000 & 1.000 (0.00) & 1.000 (0.00) & 10.00 (1.55) & 3.00 (0.77) \\
max-min & 30000 & 1.000 (0.00) & 1.000 (0.00) & 8.50 (1.02) & 2.20 (0.40) \\
\hline
inverse & 300 & 0.749 (0.21) & 0.722 (0.23) & 84.50 (68.19) & 12.10 (5.89) \\
inverse & 1000 & 0.791 (0.22) & 0.780 (0.24) & 52.30 (36.35) & 9.80 (4.56) \\
inverse & 3000 & 0.820 (0.22) & 0.798 (0.25) & 55.00 (48.09) & 9.40 (5.04) \\
inverse & 30000 & 0.969 (0.09) & 0.959 (0.12) & 37.50 (62.31) & 7.10 (6.11) \\
inverse & 300000 & 1.000 (0.00) & 1.000 (0.00) & 12.00 (0.00) & 4.00 (0.00) \\
\hline
sorted & 300 & 0.694 (0.12) & 0.618 (0.15) & 150.50 (97.52) & 29.80 (29.17) \\
sorted & 1000 & 0.872 (0.15) & 0.820 (0.19) & 162.00 (146.26) & 19.20 (20.30) \\
sorted & 3000 & 0.929 (0.14) & 0.894 (0.18) & 88.60 (121.52) & 14.20 (12.50) \\
sorted & 30000 & 0.997 (0.01) & 0.974 (0.04) & 45.80 (36.95) & 11.10 (9.03) \\
sorted & 300000 & 1.000 (0.00) & 1.000 (0.00) & 23.00 (0.00) & 8.00 (0.00) \\
\hline
\end{tabular}
\end{center}
\caption{Mean accuracy results (with standard deviation) for the top individual after 1500 steps using the evolutionary algorithm. PS stands for population size. Length of the program is the number of primitives and height is the maximum height of the program tree.}
\label{tab:evolutionary-algorithm-results}
\end{small}
\end{table}



\subsection{Large Language Model}

We have used~\textit{Qwen/Qwen2.5-Coder-32B-Instruct}~\cite{hui2024qwen2} as the LLM, which was the top open source LLM for coding tasks when we run the experiments.
We have used this model using the OpenRouter service~\footnote{OpenRouter:~\url{https://openrouter.ai}}.

We have run several experiments using the prompting methods described in the Methods section. Firstly, we show results using the LLM to describe the problem that the generated functions need to solve using the training data.
Then, we evaluate mutating the top function at each step using the LLM, in this case, the LLM can either provide an improved function with respect to the task and/or simplify that function.

\subsubsection{Seed individuals}
Initial generation of individuals using an LLM has been achieved in two steps.
In the first one, the LLM is prompted to provide a description of the problem given the training examples.
The description is used in a second prompt to generate potential individuals that could provide a function that solves the task.

Since the individuals generated by the LLM may not be syntactically correct, the prompt for individual generation is run several times until at least 30 individuals are added to the population. We have limited the total number of individuals generated by the LLM since it biases the selection towards individuals generated by the LLM and prevents effective exploration and exploitation.


Table~\ref{tab:evolutionary-algorithm-results-s} shows the results using the LLM to generate initial seed programs.
We have limited the population size to a maximum of 3,000 since performance with larger population sizes is already quite high, with probably limited chance for improvement.

It is challenging to see any effect on the~\textit{count} and~\textit{max-min} tasks, since the accuracy was already perfect for all population sizes.
We observe that for the~\textit{max-min} task, we obtain a shorter function.

Considering the more challenging tasks, we observe an improvement in both accuracy and function length (the shorter the better).
Compared to the results in table~\ref{tab:evolutionary-algorithm-results}, performance is improved when using the LLM generated individuals, but no improvement is seen for the \textit{sorted} task with population size 3,000.

\begin{table}[htpb!]
\begin{center}
\begin{tabular}{|c|r|r|r|r|r|}
\hline
 Task   & PS & Training & Testing & Length & Height \\
\hline
inverse & 300 & 0.808 (0.23) & 0.783 (0.25) & 77.70 (90.66) & 16.90 (19.98) \\
inverse & 1000 & 0.845 (0.24) & 0.837 (0.25) & 31.40 (25.63) & 7.60 (5.04) \\
inverse & 3000 & 1.000 (0.00) & 0.997 (0.01) & 26.40 (30.87) & 8.10 (9.50) \\

\hline
sorted & 300 & 0.832 (0.15) & 0.773 (0.19) & 111.60 (85.15) & 17.60 (11.52) \\
sorted & 1000 & 0.880 (0.17) & 0.850 (0.20) & 79.40 (71.17) & 15.30 (11.24) \\
sorted & 3000 & 0.921 (0.13) & 0.866 (0.18) & 95.80 (96.44) & 12.50 (7.26) \\

\hline
\end{tabular}
\end{center}
\caption{Mean accuracy results (with standard deviation) for the top individual after 1500 steps using the evolutionary algorithm with additional initial 30 programs generated by the large language model. PS stands for population size. Length of the program is the number of primitives and height is the maximum height of the program tree.}
\label{tab:evolutionary-algorithm-results-s}
\end{table} 

\subsubsection{Seed individuals and elitism mutation}

In these experiments, we have combined the seed individuals generated by the LLM with mutations of the top individual at each step by using the LLM, results are shown in table~\ref{tab:evolutionary-algorithm-results-se}.
The intention of the mutation by the LLM is two fold, either improve on the performance of the top function or reduce its length without reducing its accuracy.
Mutation and crossover do not prevent that functions generated this way contain predicates that are never traversed in the training set, which could decrease the performance on the testing set.

\begin{table}[htpb!]
\begin{center}
\begin{tabular}{|c|r|r|r|r|r|}
\hline
 Task   & PS & Training & Testing & Length & Height \\
\hline
inverse & 300 & 0.918 (0.14) & 0.910 (0.15) & 93.90 (69.08) & 24.80 (21.37) \\
inverse & 1000 & 0.950 (0.15) & 0.949 (0.15) & 14.20 (3.43) & 4.50 (0.67) \\
inverse & 3000 & 1.000 (0.00) & 1.000 (0.00) & 14.30 (1.27) & 4.70 (0.64) \\
\hline
sorted & 300 & 0.852 (0.18) & 0.832 (0.21) & 79.40 (61.06) & 14.90 (9.09) \\
sorted & 1000 & 0.896 (0.13) & 0.857 (0.17) & 88.90 (95.67) & 10.10 (6.67) \\
sorted & 3000 & 0.945 (0.11) & 0.917 (0.15) & 60.80 (50.58) & 10.80 (6.68) \\
\hline
\end{tabular}
\end{center}
\caption{Mean accuracy results (with standard deviation) for the top individual after 1500 steps using the evolutionary algorithm with additional initial 30 programs generated by the large language model and mutation of the top individual after each step. PS stands for population size. Length of the program is the number of primitives and height is the maximum height of the program tree.}
\label{tab:evolutionary-algorithm-results-se}
\end{table} 

\section{Discussion}
\subsection{Ensembling the results of multiple runs} \label{Ensembling the results of multiple runs}

Considering the results with the evolutionary algorithm, we see that large populations of individuals tend to be more stable than smaller populations.
On the other hand, larger populations sizes needs more memory and computing resources.
Increasing the number of steps (or generations) would help smaller populations to converge, but there is no guarantee on the number of steps that may be required.

In this section, we explore a different approach.
We assume that the random selection of individuals contribute to the exploration of the space of possible functions.
With smaller populations, specific areas of the function space might be explore and ensembling the identified functions from each run with a small population size might provide a set of individuals that allows considering many potential areas of the space of functions.

The benefits of this would be two fold:
1) Faster exploration provided by using smaller population sizes that are run several times.
2) Smaller populations require as well less memory, thus less resources are needed overall.

As shown in table~\ref{tab:evolutionary-algorithm-results-300-comb}, we can compare the average of the 10 runs for each setting with different population sizes, with an additional run that collects the top individual from each run, which are used as seed programs for an additional run.

As we can see, in all cases, when using the seed functions, the accuracy is largely improved. Also, the top function of this run has the shortest length compared to the average of the initial runs and compared to different population sizes, shown in table~\ref{tab:evolutionary-algorithm-results}.
The runs that were based on the LLM initial seeds manage to obtain even shorter functions, showing that the LLM is not only contributing to better performance but to generating better functions.
Examples of the best functions for each hyperparameter setting are shown in Appendix A.

\subsection{Analysis of the results}

A key contribution of the work is to demonstrate the impact of using LLMs to select better candidates for input into the evolutionary algorithm. We summarize the key results making this comparison in table \ref{tab:evolutionary-algorithm-vs-large-language-model}. The benefit of using LLMs is clearly demonstrated for the inverse problem at population sizes of 300, 1000 and 3000. For all three population sizes the LLM delivers the better solution not only on training and test metrics, but also in terms of delivering a shorter program and shallower program tree, with the tree height at 300 being an exception. 

In addition to the LLM run, we also include results in which we ensembled the results of \textit{n} runs, see section \ref{Ensembling the results of multiple runs}. This result is noted as ${LLM+}$ in table \ref{tab:evolutionary-algorithm-vs-large-language-model}. This approach delivers the best result for all population sizes, with perfect score at a population size of 3000. We also note that this delivers the smallest standard deviation on all results reported, indicating the robustness of this approach. Looking at the sorting problem, we still observe the LLM as being the clear winner at the smaller population sizes of 300 and 1000, but this is not true for the population size of 3000. Although we see high accuracy values in both the pure evolutionary algorithm and the LLM case, we note the LLM as being slightly lower in training and test accuracies. However, the ${LLM+}$ run again shows the robustness and superiority of this method as this is the top result for all population sizes. It is worth noting that in the case of the sorting task, values are much closer to each other compared to the inverse problem, which could be indicative of the complexity of the sorting vs. inverse problem. 

Thus, we draw the following conclusions: larger population sizes leads to finding a better program to solve the problems at hand, the use of an LLM reduces the search space and leads to better solutions, and finally ensembling the top programs from multiple runs presents a robust method to solving the problems.

\begin{table}[htpb!]
\begin{small}
\begin{center}
\begin{tabular}{|c|r|r|r|r|r|}
\hline
 Task   & PS  & Training & Testing & Length & Height \\
\hline
inverse$^{EA}$ & 300 & 0.749 (0.21) & 0.722 (0.23) & 84.50 (68.19) & 12.10 (5.89) \\
inverse$^{LLM}$ & 300 & 0.808 (0.23) & 0.783 (0.25) & 77.70 (90.66) & 16.90 (19.98) \\
inverse$^{LLM+}$ & 300 & 0.918 (0.14) & 0.910 (0.15) & 93.900 (69.08) & 24.800 (21.37) \\

inverse$^{EA}$ & 1000 & 0.791 (0.22) & 0.780 (0.24) & 52.30 (36.35) & 9.80 (4.56) \\
inverse$^{LLM}$ & 1000 & 0.845 (0.24) & 0.837 (0.25) & 31.40 (25.63) & 7.60 (5.04) \\
inverse$^{LLM+}$ & 1000 & 0.950 (0.15) & 0.949 (0.15) & 14.20 (3.43) & 4.50 (0.67) \\

inverse$^{EA}$ & 3000 & 0.820 (0.22) & 0.798 (0.25) & 55.00 (48.09) & 9.40 (5.04) \\
inverse$^{LLM}$ & 3000 & 1.000 (0.00) & 0.997 (0.01) & 26.40 (30.87) & 8.10 (9.50) \\
inverse$^{LLM+}$ & 3000 & 1.000 (0.00) & 1.000 (0.00) & 14.300 (1.27) & 4.700 (0.64) \\

\hline
sorted$^{EA}$ & 300 & 0.694 (0.12) & 0.618 (0.15) & 150.50 (97.52) & 29.80 (29.17) \\
sorted$^{LLM}$ & 300 & 0.832 (0.15) & 0.773 (0.19) & 111.60 (85.15) & 17.60 (11.52) \\
sorted$^{LLM+}$ & 300 & 0.852 (0.18) & 0.832 (0.21) & 79.400 (61.06) & 14.900 (9.09) \\

sorted$^{EA}$ & 1000 & 0.872 (0.15) & 0.820 (0.19) & 162.00 (146.26) & 19.20 (20.30) \\
sorted$^{LLM}$ & 1000 & 0.880 (0.17) & 0.850 (0.20) & 79.40 (71.17) & 15.30 (11.24) \\
sorted$^{LLM+}$ &1000& 0.896 (0.13) & 0.857 (0.17) & 88.90 (95.67) & 10.10 (6.67) \\

sorted$^{EA}$ & 3000 & 0.929 (0.14) & 0.894 (0.18) & 88.600 (121.52) & 14.200 (12.50) \\
sorted$^{LLM}$ & 3000 & 0.921 (0.13) & 0.866 (0.18) & 95.80 (96.44) & 12.50 (7.26) \\
sorted$^{LLM+}$ & 3000 & 0.945 (0.11) & 0.917 (0.15) & 60.800 (50.58) & 10.800 (6.68) \\
\hline
\end{tabular}
\end{center}
\caption{Mean accuracy results (with standard deviation) comparing the evolutionary algorithm, the case where the LLM was used as part of the evolutionary algorithm and LLM mutation of the top function. $^{EA}$ indicates evolutionary algorithm only, $^{LLM}$ indicates that the run was supported by seeds of individuals provided by the LLM and $^{LLM+}$ includes LLM top function mutations.}
\label{tab:evolutionary-algorithm-vs-large-language-model}
\end{small}
\end{table}








\begin{table}[htpb!]
\begin{small}
\begin{center}
\begin{tabular}{|c|r|r|r|r|r|}
\hline
 Task   & PS  & Training & Testing & Length & Height \\
\hline
inverse$^{EA}$  & 300 & 0.749 (0.21) & 0.722 (0.23) & 84.50 (68.19) & 12.10 (5.89) \\
inverse$^{EA}$* & 300 & 1.000 & 1.000 & 15 &  5 \\

inverse$^{EA}$ & 1000 & 0.791 (0.22) & 0.780 (0.24) & 52.30 (36.35) & 9.80 (4.56) \\
inverse$^{EA}$* & 1000 & 1.000 & 1.000 & 15 & 6 \\

inverse$^{EA}$ & 3000 & 0.820 (0.22) & 0.798 (0.25) & 55.00 (48.09) & 9.40 (5.04) \\
inverse$^{EA}$* & 3000 & 1.000 & 1.000 & 17 & 5 \\

\hline
sorted$^{EA}$  & 300 & 0.694 (0.12) & 0.618 (0.15) & 150.50 (97.52) & 29.80 (29.17) \\
sorted$^{EA}$* & 300 & 1.000 & 1.000 & 27 & 7 \\

sorted$^{EA}$ & 1000 & 0.872 (0.15) & 0.820 (0.19) & 162.00 (146.26) & 19.20 (20.30) \\
sorted$^{EA}$* & 1000 & 1.000 & 1.000 & 22 & 4 \\

sorted$^{EA}$ & 3000 & 0.977 (0.05) & 0.950 (0.10) & 68.20 (81.54) & 12.70 (13.23) \\

sorted$^{EA}$* & 3000 &1.000& 1.000 & 18 & 4 \\

\hline
\hline
inverse$^{LLM}$ & 300 & 0.808 (0.23) & 0.783 (0.25) & 77.70 (90.66) & 16.90 (19.98) \\
inverse$^{LLM}$* & 300 & 1.000 & 1.000 & 13 & 4 \\

inverse$^{LLM}$  & 1000 & 0.845 (0.24) & 0.837 (0.25) & 31.40 (25.63) & 7.60 (5.04) \\
inverse$^{LLM}$* & 1000 & 1.000 & 1.000 & 12 & 4 \\

inverse$^{LLM}$ & 3000 & 1.000 (0.00) & 0.997 (0.01) & 26.40 (30.87) & 8.10 (9.50) \\
inverse$^{LLM}$ & 3000 & 1.000  & 1.000 & 13 & 4 \\

\hline

sorted$^{LLM}$ & 300 & 0.832 (0.15) & 0.773 (0.19) & 111.60 (85.15) & 17.60 (11.52) \\
sorted$^{LLM}$* & 300 & 1.000 & 1.000 & 18 & 6 \\

sorted$^{LLM}$  & 1000 & 0.880 (0.17) & 0.850 (0.20) & 79.40 (71.17) & 15.30 (11.24) \\
sorted$^{LLM}$* & 1000 & 1.000 & 1.000 & 22 & 8 \\

sorted$^{LLM}$ & 3000 & 0.921 (0.13) & 0.866 (0.18) & 95.80 (96.44) & 12.50 (7.26) \\
sorted$^{LLM}$* & 3000 & 1.00 & 1.00 & 19 & 4 \\

\hline
\hline
inverse$^{LLM+}$  & 300 & 0.918 (0.14) & 0.910 (0.15) & 93.900 (69.08) & 24.800 (21.37) \\
inverse$^{LLM+}$* & 300 & 1.000 & 1.000 & 14 & 4 \\
inverse$^{LLM+}$  & 1000 & 0.950 (0.15) & 0.949 (0.15) & 14.20 (3.43) & 4.50 (0.67) \\
inverse$^{LLM+}$* & 1000 & 1.000 & 1.000 & 12 & 4 \\
inverse$^{LLM+}$  & 3000 & 1.000 (0.00) & 1.000 (0.00) & 14.300 (1.27) & 4.700 (0.64) \\
inverse$^{LLM+}$* & 3000 & 1.000 & 1.000 & 12 & 4 \\
\hline
sorted$^{LLM+}$  & 300  & 0.852 (0.18) & 0.832 (0.21) & 79.400 (61.06) & 14.900 (9.09) \\
sorted$^{LLM+}$* & 300  & 1.000 & 1.000 & 15 & 5 \\
sorted$^{LLM+}$  & 1000 & 0.896 (0.13) & 0.857 (0.17) & 88.90 (95.67) & 10.10 (6.67) \\
sorted$^{LLM+}$* & 1000 & 1.000 & 1.000 & 15 & 5 \\
sorted$^{LLM+}$  & 3000 & 0.945 (0.11) & 0.917 (0.15) & 60.800 (50.58) & 10.800 (6.68) \\
sorted$^{LLM+}$* & 3000 & 1.000 & 1.000 & 21 & 6 \\
\hline
\end{tabular}
\end{center}
\caption{Mean accuracy results (with standard deviation) for the top individual after 1500 steps using the evolutionary algorithm. PS stands for population size. Length of the program is the number of primitives and height is the maximum height of the program tree. $^{EA}$ indicates evolutionary algorithm only, $^{LLM}$ indicates that the run was supported by seeds of individuals provided by the LLM and $^{LLM+}$ includes LLM top function mutations. * indicates that it is a run combining the top individuals as seeds.}
\label{tab:evolutionary-algorithm-results-300-comb}
\end{small}
\end{table}

\section{Conclusion and Future Work}

Deep learning has made massive strides in being able to solve problems with human or superhuman accuracy. However, these problems require a huge volume of training data allowing the model to learn the distribution of the data and it is able to make predictions. We presented an alternative approach, in which we allow the model to search for the right solution. We provide some primitives that we believed is required to solve a problem domain and ask the model to synthesize a program that will be able to solve the problem at hand. We employed EAs to find the solution and supplemented it with LLMs to narrow down the vast search space of possible solutions. To address the computational bottleneck of evaluating large populations, we introduced a fast evaluation framework capable of rapid fitness assessment. This framework is compatible with both CPU and GPU, utilizing CUDA. 

Our findings indicate the benefit of using LLMs in this context and we showed that ensembling the results of \textit{n} runs leads to more accurate and robust solutions. For the problem of list inversion, our approach delivers a function that can solve all training and test cases in our data with perfect score. For list sorting, we obtained an average training accuracy of 94.5\% and a test accuracy of 91.7\%. Also, we presented an ensemble approach that manages to solve all the problems, even with small population sizes. Thus, by taking smaller samples of the search space and combining them reduces the need for very large population sizes and allows for an optimal solution to be found in a reasonable time frame. Although we show the strengths of this approach to synthesize programs that can solve problems such as list inversion and sorting, we acknowledge the limitations of the problem space explored. Solving more complex problems is left as future work.


\bibliographystyle{plain}
\bibliography{bibliography}

\newpage
\section*{Appendix A - Example functions generated by the best hyperparameter setting}

\begin{listing}[htpb!]
\begin{minted}[frame=lines, mathescape, fontsize=\footnotesize]{python}
testing_output_write(get_testing_length_input_x())
\end{minted}
\caption{Count task learned example function}
\label{lst:count_code}
\end{listing}

\begin{listing}[htpb!]
\begin{minted}[frame=lines, mathescape, fontsize=\footnotesize]{python}
comparison(
    bigger_thanW(input_max(), output_read()),
    testing_output_write(testing_input_min()),
    testing_output_write(testing_input_max())
)
\end{minted}
\caption{Max-min task learned function}
\label{lst:max_min_code}
\end{listing}

\begin{listing}[htpb!]
\begin{minted}[frame=lines, mathescape, fontsize=\footnotesize]{python}
prog2(
    loop(
        get_testing_length_input_x(),
        testing_input_move_right()
    ),
    loop(
        get_testing_length_input_x(),
        prog2(
            testing_output_write(testing_input_read()),
            prog2(
                testing_input_move_left(),
                testing_output_move_right()
            )
        )
    )
)
\end{minted}
\caption{Inverse task learned example function}
\label{lst:inverse_code}
\end{listing}

\begin{listing}[htpb!]
\begin{minted}[frame=lines, mathescape, fontsize=\footnotesize]{python}
prog2(
    testing_reset_output_position(), 
    loop(
        get_testing_length_input_x(), 
        prog2(
            prog2(
                comparison(
                    bigger_than_testing_output_next(),
                    swap_testing_output_next(),
                    testing_input_move_right()
                ),
                comparison(
                    bigger_than_output_next(),
                    swap_testing_output_next(),
                    output_move_right()
                )
            ),
            testing_output_move_right()
        )
    )
)
\end{minted}
\caption{Sorting task learned example function}
\label{lst:sorted_code}
\end{listing}

\end{document}